\pdfoutput=1

\documentclass[11pt]{article}

\usepackage[preprint]{acl}

\usepackage{times}
\usepackage{latexsym}

\usepackage[T1]{fontenc}

\usepackage[utf8]{inputenc}

\usepackage{microtype}

\usepackage{inconsolata}

\usepackage{graphicx}

\usepackage{booktabs}
\usepackage{tabularx}
\usepackage{pdfpages}
\usepackage[justification=centering]{caption}
\usepackage{svg}
\usepackage{amsmath}

\title{Shiksha: A Technical Domain focused Translation Dataset and Model for Indian Languages}

\author{
    Advait Joglekar \and S. Umesh \\
    SPRING Lab, \\
    Indian Institute of Technology Madras, India \\
    \texttt{\href{mailto:advaitjoglekar@gmail.com}{advaitjoglekar@gmail.com}}, 
    \texttt{\href{mailto:umeshs@ee.iitm.ac.in}{umeshs@ee.iitm.ac.in}}
}

\begin{document}

\maketitle

\begin{abstract}

\textit{Neural Machine Translation} (NMT) models are typically trained on datasets with limited exposure to \textbf{Scientific, Technical and Educational domains}. Translation models thus, in general, struggle with tasks that involve scientific understanding or technical jargon. Their performance is found to be even worse for \textbf{low-resource Indian languages}. Finding a translation dataset that tends to these domains in particular, poses a difficult challenge. In this paper, we address this by creating a multilingual parallel corpus containing \textbf{more than 2.8 million rows} of \textit{English-to-Indic} and \textit{Indic-to-Indic} high-quality translation pairs across \textbf{8 Indian languages}. We achieve this by bitext mining \textbf{human-translated transcriptions of NPTEL\footnote{\href{https://nptel.ac.in}{https://nptel.ac.in}} video lectures}. We also finetune and evaluate NMT models using this corpus and surpass all other publicly available models at in-domain tasks. We also demonstrate the potential for generalizing to out-of-domain translation tasks by improving the baseline by over 2 BLEU on average for these Indian languages on the Flores+ benchmark. We are pleased to release our model and dataset via this link: \href{https://huggingface.co/SPRINGLab}{https://huggingface.co/SPRINGLab}. 
\end{abstract}

\section{Introduction}

NPTEL (\textit{National Programme on Technology Enhanced Learning}) has long been a valuable resource for free on-demand higher-educational content across a diverse range of specialized disciplines. Over the past two decades since its inception, NPTEL has curated an extensive library of over 56,000 hours of video lectures, all made publicly available along with their audio transcriptions in an easily accessible manner. In response to the growing number of Indian students, NPTEL has taken steps to support Indian language transcriptions for more than 12,000 hours of video content. These captions are primarily translations of the original English transcriptions, carefully crafted by subject-matter experts. This multi-year endeavor has led to the creation of a high-quality parallel textual resource spanning multiple Indian languages, covering various fields in the Scientific, Engineering, and Humanities domains. Our research leverages this rich data resource to develop competitive Machine Translation (MT) models specifically tailored for Indian languages. Additionally, we investigate how models fine-tuned on this data can assist human translators and help accelerate the mission of providing accurate Indic subtitles for all NPTEL video lectures. This effort aims to benefit a large audience of Indian students struggling with the lack of university-level educational content in their native tongues.

\begin{figure}
    \centering
    \includegraphics[width=1\linewidth]{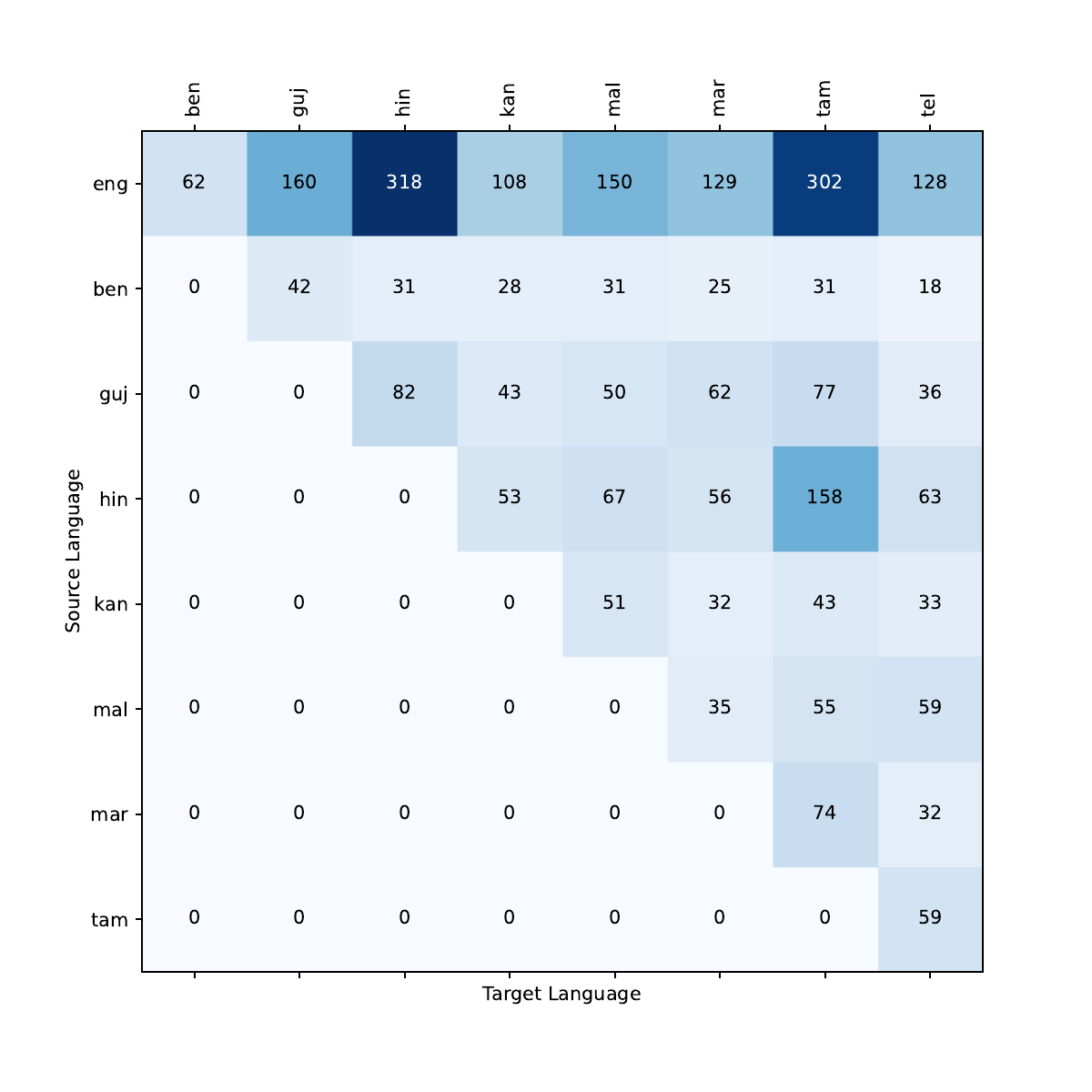}
    \caption{Translation Pair Counts (in thousands)}
    \label{fig:pair-counts}
\end{figure}

\begin{table*}[t]
    \centering
    \includegraphics[width=1\linewidth]{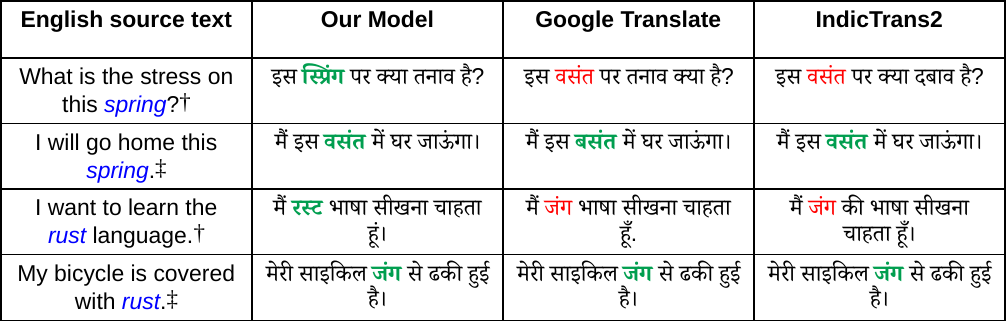}
    \caption{
        Example translations from English to Hindi in the Scientific/Technical domain. \\
        Sentences marked as † are in-domain, while ‡ are out-of-domain. \\
        The words in \textit{blue} are terms with multiple meanings, that tend to get translated incorrectly. \\
        The words in \textbf{green} represent the correct, expected translation by the model for the \textit{blue} word in the given context. The words in \underline{red} represent incorrect translations. 
        }
    \label{tab:comparison-table}
\end{table*}

\section{Where does Present-day MT fail?}

Let's quickly look at how Machine Translation models in use widely today perform on Technical-domain tasks and instances in which they fail. 

Consider the text "I want to learn the rust language." Here we are talking about the programming language Rust and not the chemical phenomena. From Table \ref{tab:comparison-table} we can see that both Google Translate~\footnote{\href{https://translate.google.com/}{https://translate.google.com/}} and IndicTrans2 (IT2) \cite{gala2023indictrans} get their Hindi translations wrong. Not only that, if we backtranslate their results we get the sentence: "I want to learn the language of war," which is very far from what we originally meant. This happens in this case because the Hind word for "rust" has two meanings; the phenomena of rust and also war. Thus, in such situations it is very important for the translation model to understand the context well since the meaning of a sentence can completely change with the wrong choice of a word. So, with this we can see that current models are prone to making mistakes for tasks in these domains. Our paper hopes to alleviate such shortcomings.



\section{Related Work}
In related work like Samanantar \cite{ramesh-etal-2022-samanantar} and IndicTrans2 \cite{gala2023indictrans}, NPTEL has been identified as a useful resource for Machine Translation (MT). These two studies in particular attempt at mining for parallel sentence pairs by utilizing various sources on the internet, including this one. Due to the lack of precise information given in these papers, we are unable to know the exact quantity of sentence-pairs mined from NPTEL with certainty. Regardless of this, we have found some key issues with their data. A significant quantity of extracted sentence-pairs was found to be composed of unfiltered artifacts like timestamps. Several instances of code-mixed sentences have also been miscategorized as English, leading to poor alignment quality and under-exploitation of raw data. Their alignments too were limited to 1-1 sentence matchings, leaving us room for better alignments with n-m translation pairs. The data they mined was solely in the English-Indic direction as well, despite there being a significant potential for mining Indic-Indic translation pairs from this source. In this work, we attempt to alleviate these shortcomings.

\begin{figure*}
    \centering
    \includegraphics[width=1\linewidth]{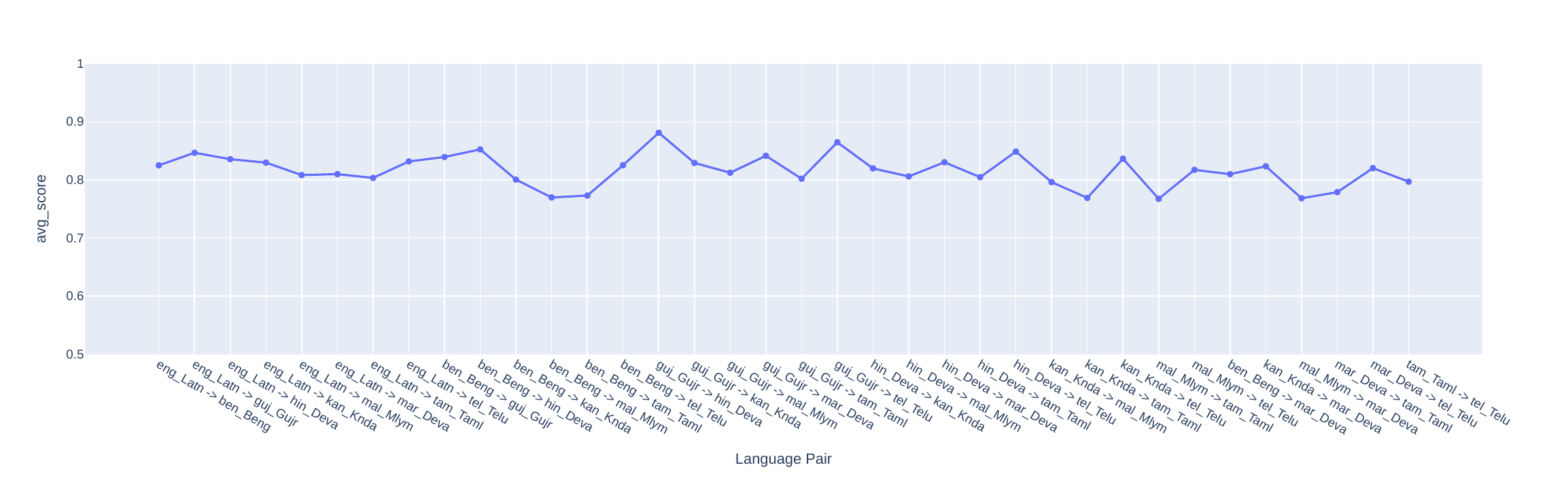}
    \caption{Average LABSE score across language pairs}
    \label{fig:avg-scores}
\end{figure*}

\section{The Dataset}

\subsection{First, the Source}

The initial step in creating any dataset involves obtaining the raw data. Instead of scraping subtitles from YouTube videos, we obtained the raw data from NPTEL. We were provided a list of 10,669 videos and their corresponding transcriptions and related metadata. These transcriptions were bilingual documents spanning 8 languages~\footnote{Bengali, Gujarati, Hindi, Kannada, Malayalam, Marathi, Tamil and Telugu}, featuring alternating English and Indic text, interspersed with reference timestamps and video snapshots. Refer to Appendix \ref{appendix:source-page} for a sample page.

\subsection{Data Cleaning and Extraction}

Given the unusual format of these documents, we wrote a Python script to extract the meaningful text data from it while avoiding any timestamp references. In this script, we first pull out the text from these documents and then use regex patterns to filter out the timestamps. We then used simple paragraph segmenting tools from \textit{nltk} \cite{nltk} and \textit{indic-nlp}~\cite{kunchukuttan2020indicnlp} libraries to identify and separate English and Indic sentences. With this, the lectures are now decomposed into parallel documents of text stored in separate files to create a massive bitext corpora. 

\subsection{Bitext Mining}

With this parallel corpora in place, we begin the most crucial part: Bitext mining. Our objective is to find as many sentence-pairs as we can from the source data while still maintaining high confidence in their translation accuracy. Luckily, Sentence-Alignment is a well studied problem dating as far back as \citeyear{brown-etal-1991-aligning} (\citeauthor{brown-etal-1991-aligning}).

Recent work like Vecalign \cite{thompson-koehn-2019-vecalign} has focused on using multi-lingual embedding models to find pairs based on vector similarity of sentence embeddings. These have been shown to achieve state-of-the-art performance, significantly surpassing previous approaches. In our work, we use SentAlign \citep{steingrimsson-etal-2023-sentalign} which employs LABSE \citep{feng-etal-2022-language} along with optimized alignment algorithms to mine parallel documents with high accuracy and efficiency. With this we are also able to find 1-n and n-1 sentence matches. 

\subsection{Data Collation}
Each lecture now has its own documents with all its sentences aligned into bilingual sentence-pairs. We collect these pairs and combine them, along with their lecture metadata, into a massive translation dataset. After post-processing with deduplication, we arrive at a corpus of roughly 2.8 million sentence-pairs.

\subsection{Data Analysis} 

To understand the quality and quantity of this data, we must first thoroughly analyse it. Our dataset has 8 English-Indic and 28 Indic-Indic language pairs. This means that there exists at least one common set of lectures among each language-pair, providing us with inter-Indic alignments for all languages covered in this dataset. We find 48.6\verb|%| of these to be English-to-Indic language pairs. This is the direct result of English lectures having been translated into multiple languages, albeit with arbitrary combinations, giving us a robust Indic-to-Indic data subset. 

For assessing the alignment quality of our translation pairs, we look at the average LABSE similarity scores as the primary measure. The plot of this metric (Figure~\ref{fig:avg-scores}) demonstrates a strong consistency in scores across all languages despite differences in the quantity of the mined sentence-pairs. These data points are also seen to be tending towards 0.8 and are never below 0.75, validating our confidence in the quality of the source data and the accuracy of our alignments.

\begin{table*}[ht]
    \centering
    \begin{tabular}{lcc}
        \multicolumn{2}{c}{\textbf{Our testset}} \\
        \hline
        \toprule
        \textbf{Models}  & \textbf{en-in}\\ \midrule
        NLLB    & 30.73 / 57.62 \\
        LoRA FT & \textbf{48.98 / 71.99} \\
        IT2     & 39.66 / 66.49 \\
        \bottomrule
    \end{tabular}
    \hspace{40pt}
    \begin{tabular}{lcc}
        \multicolumn{2}{c}{\textbf{Flores+}} \\
        \hline
        \toprule
        \textbf{Models}  & \textbf{en-in}\\ \midrule
        NLLB    & 19.73 / 54.27 \\
        LoRA FT & 22.04 / 57.33 \\
        IT2     & \textbf{24.08 / 59.45} \\ 
        \bottomrule
    \end{tabular}
    \caption{Results are in the form <bleu>/<chrf++>. \\ These scores represent the average of all 8 languages covered by the dataset. \\ All models were evaluated without using beam-search or sampling.}
    \label{tab:flores}
\end{table*}

\section{The Model}

The value of a dataset can only be best quantified when a model has been trained with it, evaluated against others and the results analyzed. We wish to fine-tune and evaluate a powerful MT base model to test the hypothesis that our dataset can help improve the performance of translation tasks in the Technical domain. We will test if existing models can be improved meaningfully by being fine-tuned on our dataset.

\subsection{Baseline Model selection}

When it comes to choosing a strong multi-lingual model that is, or at least close to, the state-of-the-art, we find that our options are limited. IndicTrans2 could be a good choice, except that it provides different models for English-Indic, Indic-English and Indic-Indic directions. We eliminate this option since we hope to leverage transfer learning by training just one model in all the 36 language-pair combinations that our dataset supports. That leaves us with only two possible candidates: NLLB-200 \cite{nllbteam2022language} and MADLAD-400 \cite{kudugunta2023madlad400}. These are both massively multilingual Transformer models that promise to be the right baseline for our study. We choose NLLB on the basis of its superior evaluation scores for Indian languages on the Flores-200 \cite{goyal-etal-2022-flores} benchmark, according to the results published in the MADLAD paper.

\subsection{Training}

NLLB-200 models are available in a wide variety of sizes ranging from a 600M parameter distilled model to a massive 54B parameter Mixture-of-Experts model. For our experiments, we decide to choose the 3.3B parameter version as a sweet spot between performance quality and compute requirements. 

Still, even with this relatively smaller NLLB 3.3B model, running a Full Fine-Tuning (FFT) setup can turn out to be a very compute intensive endeavour. The amount of time required to effectively train our model will also be significant. In our case we wish to execute a number of experiments with different approaches in hopes to achieve the best results. FFT thus would not be a feasible approach. Instead, we decide to utilize a Parameter-Efficient Fine Tuning (PEFT) method known as Low-Rank Adaptation (LoRA) \cite{hu2022lora} to train our model.

We primarily trained three models using three different approaches. All of them were done using LoRA with NLLB 3.3B. These approaches included: 1) training a model purely on our dataset in one direction, 2) training using Curriculum Learning (CL) \cite{10.1145/1553374.1553380} with a cleaned subset of the BPCC corpus \cite{gala2023indictrans} with our 8 Indian languages, comprising of 4 million rows,  before introducing our dataset, 3) training on a massive 12 million samples which included the cleaned BPCC corpus and our dataset in both directions. All our models were trained on a node of 8 NVIDIA A100 40GB GPUs. Evaluation results for all the three models were found to be similar, with our 3rd approach performing slightly better. The hyperparameters and detailed results for all three are available in Appendix \ref{appendix:model}.

\subsection{Evaluation}

For evaluation we compare our third model, trained on 12 million rows, with the baseline NLLB model and the 1B parameter IndicTrans2 model. For an in-domain test, we used the top one thousand rows (by LABSE score) of our held-out test set for each language. Our model outperforms the rest on our test set and demonstrates the efficacy of our model at translations involving the technical domain. We further test our models on the Flores+ \footnote{\href{https://github.com/openlanguagedata/flores}{https://github.com/openlanguagedata/flores}} (Previously Flores-200) devtest set. We find that our model is also able to generalize well, as seen from the improvements on the baseline scores. Our results manage to come closer to IndicTrans2, which was trained on a corpus far larger than ours. These scores are depicted in Table \ref{tab:flores} above. Language-wise comparison of evaluation scores are also available in Appendix \ref{appendix:model}.

\section{Translingua}

This research goes beyond just experiments. Our models are now built into a tool called Translingua, that is being widely used by human annotators across India to translate NPTEL lecture transcripts into more languages than ever before, with far better speed and accuracy.  A screenshot of this tool along with feedback of the users on translation quality is available in Appendix \ref{appendix:translingua}. 

\section{Conclusion}

In this paper, we introduced Shiksha, a novel translation dataset and model tailored for Indian languages, with a particular focus on the Scientific, Technical, and Educational domains. We created a robust multilingual parallel corpus consisting of over 2.8 million high-quality translation pairs across 8 Indian languages. Our approach involved meticulous data extraction, cleaning, and bitext mining to ensure the accuracy and relevance of the dataset. We also fine-tuned state-of-the-art baseline NMT models using this dataset and demonstrated significant performance improvements in not only in-domain, but also out-of-domain translation tasks. 

With this paper, we wish to encourage the importance of domain-specific datasets in advancing NMT capabilities. We believe that our dataset and models will serve as valuable resources for the community and foster further research in multilingual NMT.

\section{Limitations}

Despite the promising results of our dataset and model, there are some limitations that need to be acknowledged:

\begin{itemize}
    \item The dataset is heavily skewed towards scientific, technical, and educational domains, sourced primarily from NPTEL video lectures. This can lead to degradation in translation quality for general tasks in unexpected ways that standard benchmarks may not catch. We recommend supplementing our dataset with additional diverse and balanced sources covering a wide range of domains, including everyday conversational language, literature, social media, and news articles. This will help ensure a more stable training and evaluation process, ultimately enhancing the translation system's robustness and accuracy across different contexts.

    \item We have not meaningfully tested our model's performance on Indic-English or Indic-Indic directions as our research was focused primarily on translating out of English. Our models thus may not perform well on those language directions.

    \item The quality of our translation dataset and models is heavily dependent upon the accuracy of the original NPTEL transcriptions. Any errors or inconsistencies in them are propagated into our dataset, potentially affecting the training and evaluation of the translation models. Further human evaluation might be needed to verify the quality of these translations.
\end{itemize}
\bibliography{anthology, custom}

\newpage

\section*{Appendices}
\appendix

\section{Source Document}
\label{appendix:source-page}

\begin{figure}[ht]
    \centering
    \includegraphics[width=1\linewidth]{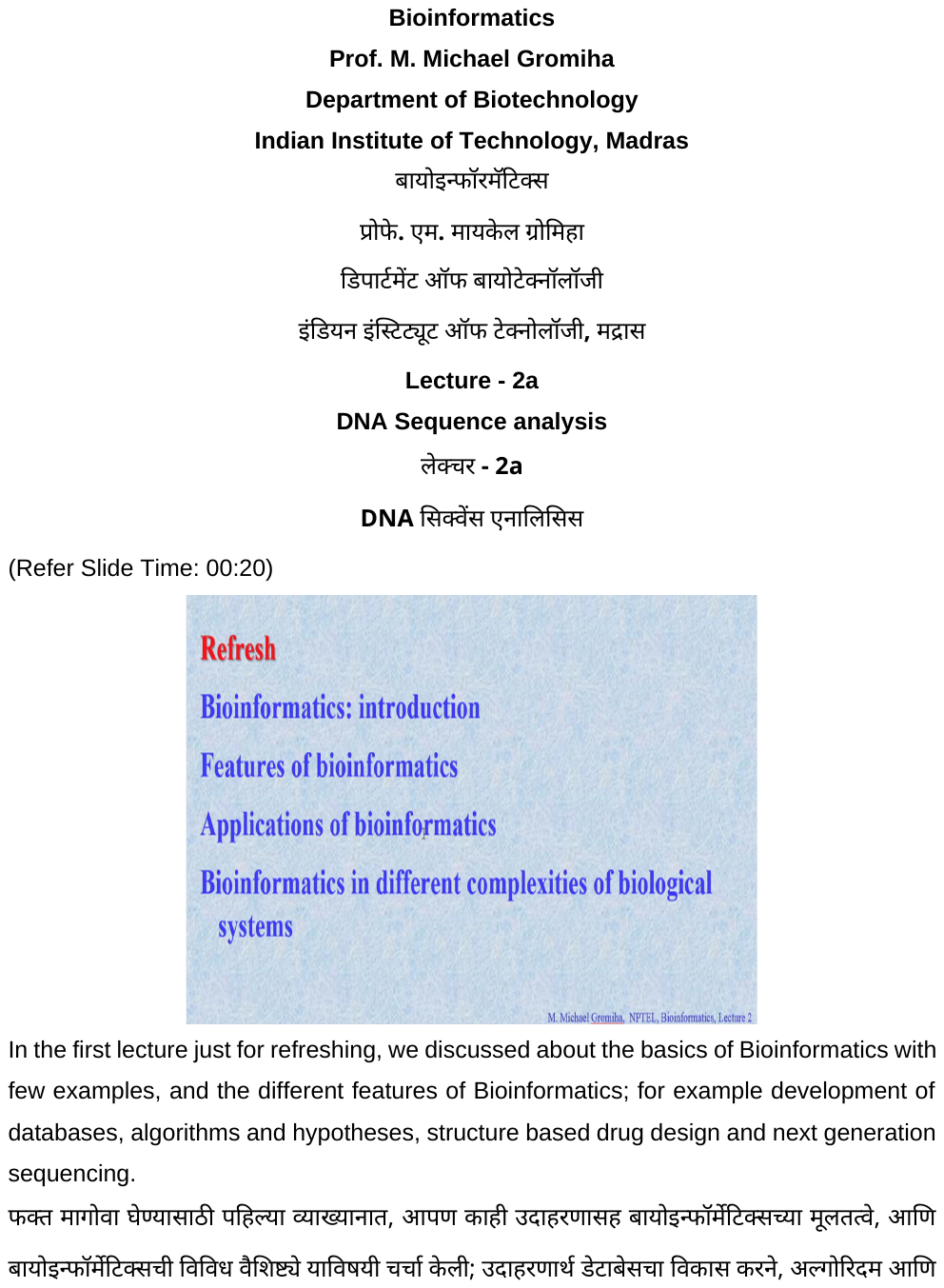}
    \caption{A sample page from a bilingual document}
    \label{figure:sample-page}
\end{figure}

\newpage

\section{Model Hyperparameters and Results}
\label{appendix:model}

\begin{table}[h]
    \centering
    \begin{tabular}{cc}
        Parameter & Setting \\ \toprule
        peft-type & LORA \\
        rank & 256 \\
        lora alpha & 256 \\
        lora dropout & 0.1 \\
        rslora & True \\
        target modules & all-linear \\
        learning rate & 4e-5 \\
        optimizer & adafactor \\
        data-type & BF-16 \\
        epochs & 1 \\ \bottomrule
    \end{tabular}
    \caption{Hyperparameters for our 3rd approach. \\ First approach was trained for 10 epochs and second for 4 epochs seperately}
    \label{tab:hyperparameters}
\end{table}

\begin{figure}[ht]
    \centering
    \includegraphics[width=1\linewidth]{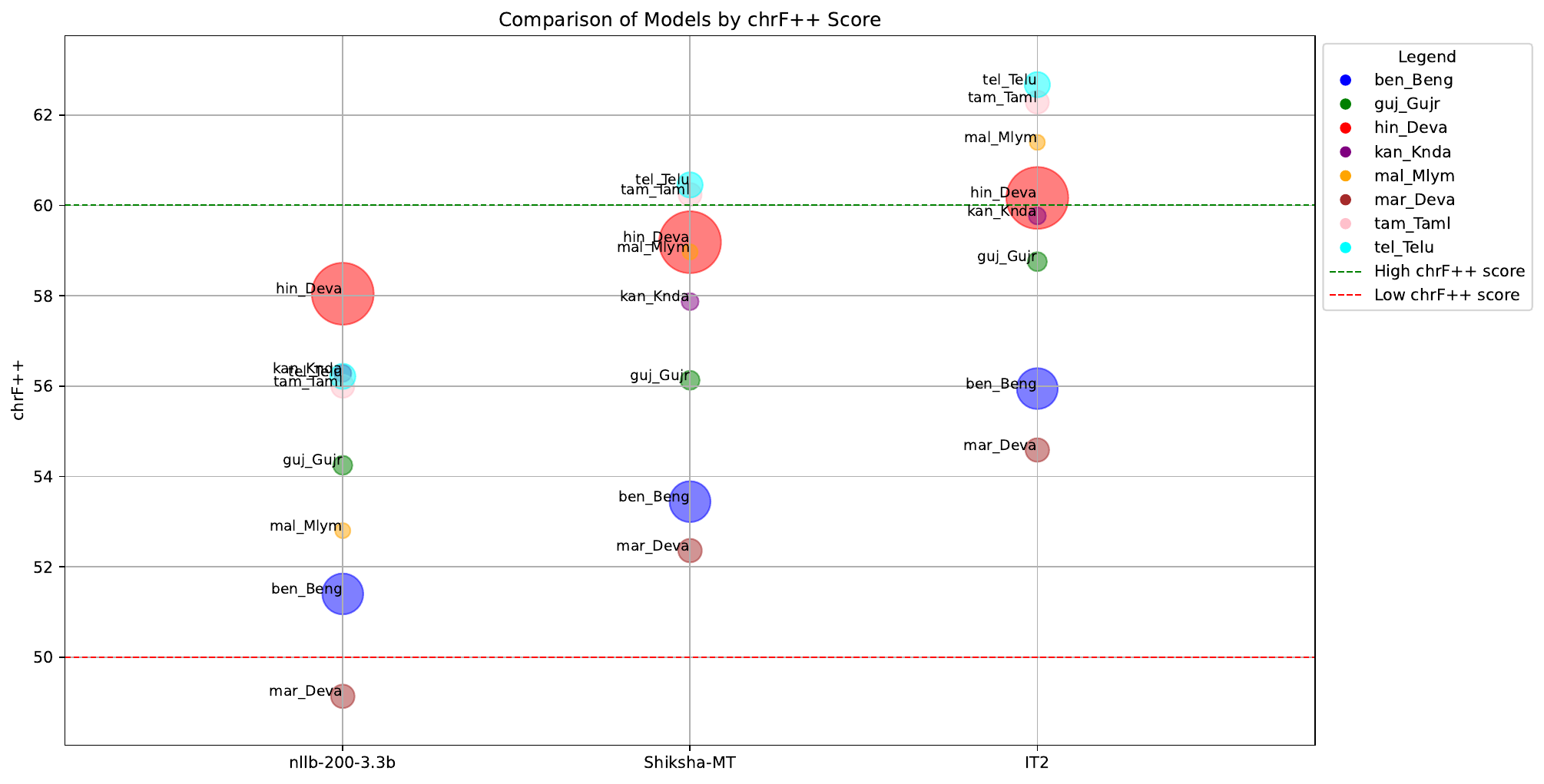}
    \caption{Chrf++ comparison between NLLB, IT2 and our model across all Indian languages. \\
        The size of the bubble represents the population of the speakers.    
        }
    \label{fig:enter-label}
\end{figure}

\section{Translingua}
\label{appendix:translingua}

\begin{figure*}[ht]
    \centering
    \includegraphics[width=1\linewidth]{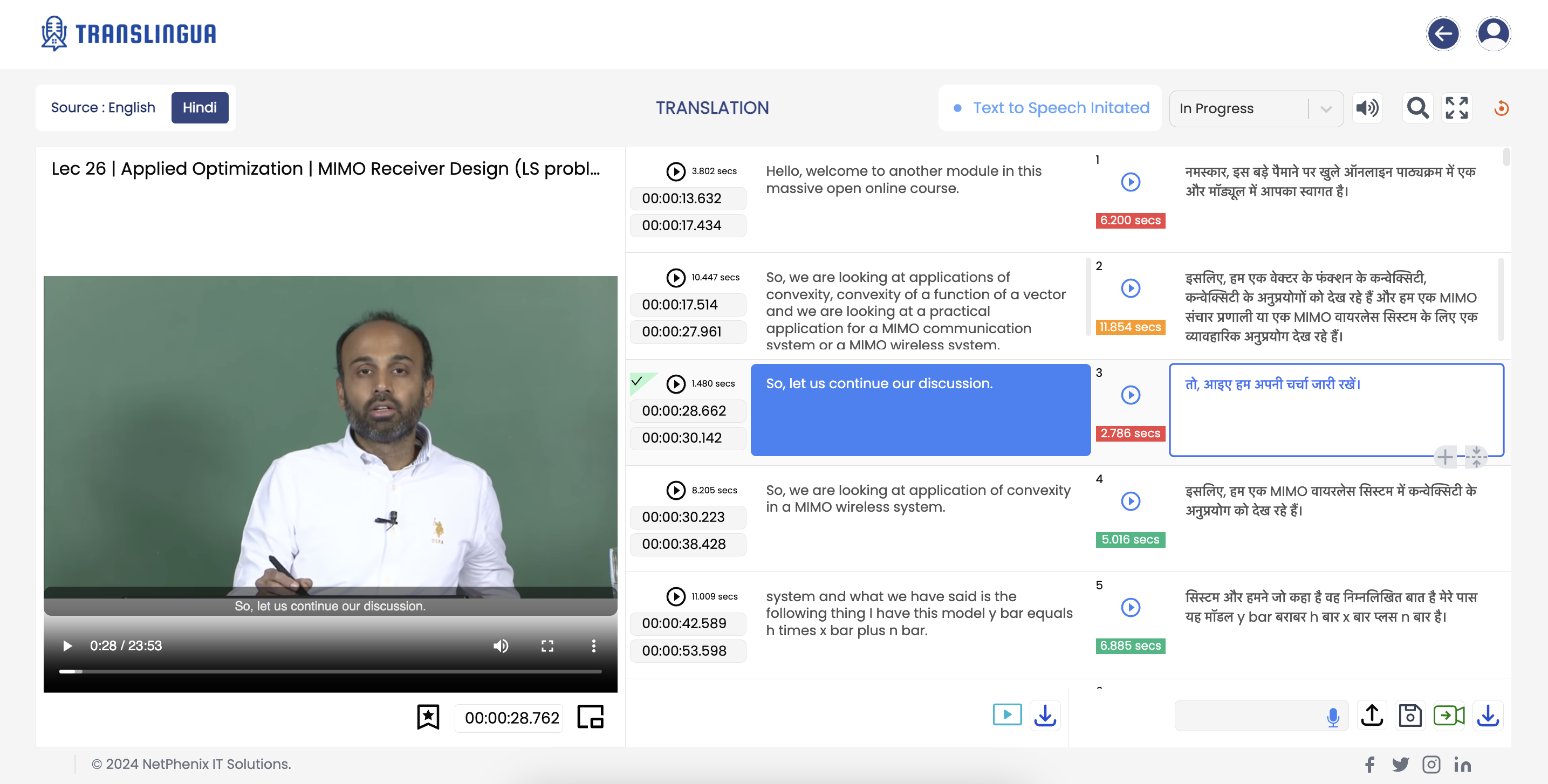}
    \caption{A screenshot from the Translingua tool}
    \label{fig:screenshot}
\end{figure*}

\begin{figure*}[hb]
    \centering
    \includegraphics[width=1\linewidth]{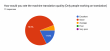}
    \caption{Feedback on Translation Quality from a subset of Users}
    \label{fig:feedback}
\end{figure*}

\end{document}